%% file: main.tex
\documentclass[runningheads]{llncs}

\usepackage{eccv}

\usepackage{eccvabbrv}

\usepackage{graphicx}
\usepackage{booktabs}

\usepackage[accsupp]{axessibility}  

\usepackage{hyperref}

\usepackage{orcidlink}

\usepackage{graphicx}
\usepackage{amsmath}
\usepackage{amssymb}
\usepackage{booktabs}
\usepackage{multirow}
\usepackage{bm}
\usepackage{bbding}

\usepackage{wrapfig} 

\usepackage[dvipsnames]{xcolor}

\usepackage{soul}

\usepackage{colortbl}

\newcommand{\ourMethod}{SEED}

\newcommand\blfootnote[1]{%
  \begingroup
  \renewcommand\thefootnote{}\footnotetext{#1}%
  \addtocounter{footnote}{-1}%
  \endgroup
}

\usepackage[capitalize]{cleveref}
\crefname{section}{Sec.}{Secs.}
\Crefname{section}{Section}{Sections}
\Crefname{table}{Table}{Tables}
\crefname{table}{Tab.}{Tabs.}

\begin{document}

\title{{\ourMethod}: A Simple and Effective 3D DETR in Point Clouds}

\titlerunning{A Simple and Effective 3D DETR in Point Clouds}

\author{Zhe Liu\inst{1*}
\and Jinghua Hou\inst{1*}
\and Xiaoqing Ye\inst{2}
\and Tong Wang\inst{2}
\and Jingdong Wang\inst{2}
\and Xiang Bai\inst{1\dag}}

\authorrunning{Z.~Liu et al.}

\institute{Huazhong University of Science and Technology
\email{\{zheliu1994,jhhou,xbai\}@hust.edu.cn}\\
\and Baidu Inc., China\\
\email{yxq@whu.edu.cn,wangtong16@baidu.com,wangjingdong@outlook.com}}

\maketitle

\blfootnote{
\noindent $*$ Equal contribution. \\
$\dag$ Corresponding author.
}

\begin{abstract}
Recently, detection transformers~(DETRs) have gradually taken a dominant position in 2D detection thanks to their elegant framework. However, DETR-based detectors for 3D point clouds are still difficult to achieve satisfactory performance. We argue that the main challenges are twofold: 1)~How to obtain the appropriate object queries is challenging due to the high sparsity and uneven distribution of point clouds; 2)~How to implement an effective query interaction by exploiting the rich geometric structure of point clouds is not fully explored.
To this end, we propose a \textbf{S}imple and \textbf{E}ff\textbf{E}ctive 3D \textbf{D}ETR method~(\textbf{\ourMethod}) for detecting 3D objects from point clouds, which involves a dual query selection~(DQS) module and a deformable grid attention~(DGA) module. More concretely, to obtain appropriate queries, DQS first ensures a high recall to retain a large number of queries by the predicted confidence scores and then further picks out high-quality queries according to the estimated quality scores. DGA uniformly divides each reference box into grids as the reference points and then utilizes the predicted offsets to achieve a flexible receptive field, allowing the network to focus on relevant regions and capture more informative features. Extensive ablation studies on DQS and DGA demonstrate its effectiveness. Furthermore, our {\ourMethod} achieves state-of-the-art detection performance on both the large-scale Waymo and nuScenes datasets, illustrating the superiority of our proposed method. \textit{The code is available at}  \url{https://github.com/happinesslz/SEED}.
\end{abstract}

\input{contents/1_intro}

\input{contents/2_related_work}
\input{contents/3_method}
\input{contents/4_exps}
\input{contents/5_conclusion}


%
%
\bibliographystyle{splncs04}
\bibliography{main}

\clearpage
\input{contents/6_supp}

\end{document}

%% file: contents/1_intro.tex
\section{Introduction}
\label{sec:intro}



DEtection TRansformer~(DETR)~\cite{carion2020end} is the pioneering end-to-end transformer-based detector, which redefines object detection as a set prediction problem and eliminates hand-crafted anchors and non-maximum suppression~(NMS) post-processing. These superior characteristics make the DETR paradigm~\cite{carion2020end,li2022dn,liu2022dab,zhang2022dino} become the mainstream method for 2D object detection tasks.

However, although many efforts have been made
towards the DETR paradigm for 3D object detection~\cite{bai2022transfusion,misra2021end,nguyen2022boxer,chen2023focalformer3d,yan2023cross,zhu2023conquer,wang2023uni3detr} in point clouds, they have not demonstrated stunning performance similar to the 2D domain and still fall behind state-of-the-art 3D detectors~\cite{wang2023dsvt,fan2022fully,shi2021pv,deng2021voxel}. The main reason is the huge gap between 2D images and 3D points~(\textit{i.e.}, dense and regular 2D images \textit{v.s.} sparse and irregular 3D points clouds), which requires us to carry out special designs for two critical components~(\textit{i.e.}, query selection and query interaction) in the DETR paradigm. For query selection, some methods~\cite{bai2022transfusion,chen2023focalformer3d,zhu2023conquer} mainly select Top-N~(\textit{e.g.}, N=200, 300 or 1000) features as queries from the score map. Although effective, these methods do not consider the quality of the selected queries for box localization. For query interaction, some works~\cite{bai2022transfusion,yan2023cross} perform several attention operations to achieve sufficient feature interaction. However, these approaches do not sufficiently take advantage of geometric information of 3D objects from point clouds.

\begin{wrapfigure}{r}{0.5\textwidth}
	\vspace{-20pt}
	\centering
	\includegraphics[width=\linewidth]{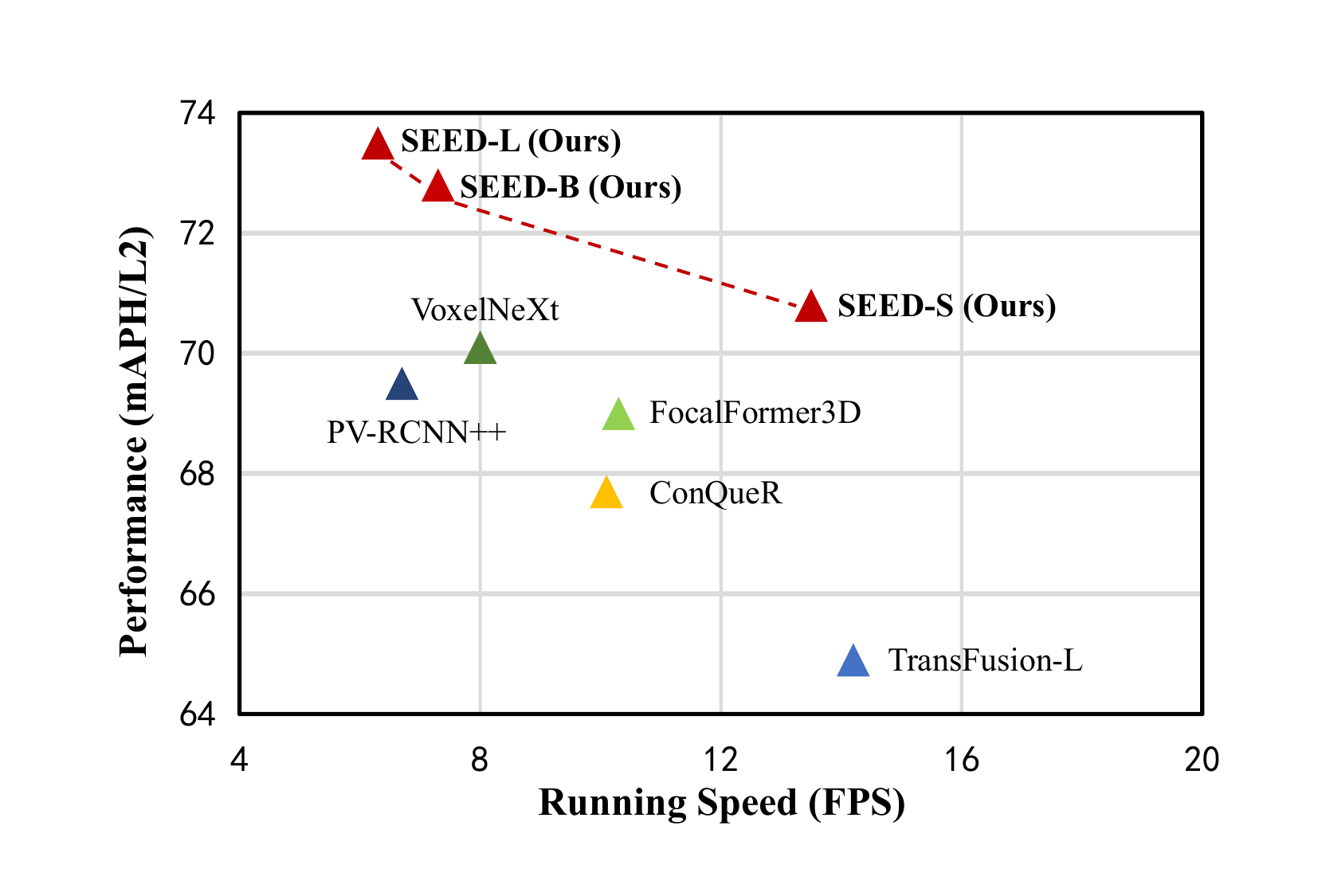}
	\vspace{-15pt}
	\caption{{
		Comparison with DETR-based detectors~\cite{zhu2023conquer,chen2023focalformer3d,bai2022transfusion} and other representative methods~\cite{chen2023voxelnext,shi2021pv} on Waymo \emph{validation} dataset~\cite{sun2020scalability} in terms of detection performance and running speed. For a fair comparison, we evaluate the running speed of all approaches on an NVIDIA GeForce RTX 3090 with a batch size of 1. \textit{-S}, \textit{-B} and \textit{-L} means the small, base, large versions of our {\ourMethod}, respectively.
		}} 
  \label{fig_intro}
	\vspace{-12pt}
\end{wrapfigure}

In this paper, to alleviate the above challenges, we propose a \textbf{S}imple and \textbf{E}ff\textbf{E}ctive 3D \textbf{D}ETR method~(\textbf{\ourMethod}) for detecting 3D objects from point clouds.
The first key design in our {\ourMethod} is the proposed dual query selection~(DQS) to pick out high-quality queries in a coarse-to-fine manner, which includes a foreground query selection and a quality query selection.
This manner is different from existing methods by one-step query selection~\cite{bai2022transfusion,zhu2023conquer}. More concretely, to ensure a high recall, we first retain a large number of foreground queries in the foreground query selection according to the estimated confidence scores from a mask predictor. Then, we employ a {\ourMethod} decoder layer to allow these queries to effectively interact with Bird's Eye View~(BEV) features. The enhanced queries are fed into the stage of quality query selection to pick out high-quality queries.

The second core design in our {\ourMethod} is the proposed deformable grid attention~(DGA) to make the network focus on relevant regions and achieve more effective feature interaction. Specifically, to exploit the rich geometric information in point clouds, we first divide the referenced box estimated by a regression branch into uniform grids, whose corresponding features can be easily collected to describe the geometric structures of 3D objects. To alleviate the strong dependence on the high-accuracy reference box, we further use these sampling grids as reference points and apply the predicted offsets to obtain flexible receptive fields. This enables the network to focus on surrounding regions of interest, even for less precise reference boxes.

As shown in Figure~\ref{fig_intro}, we compare our {\ourMethod} with the existing DETR-based 3D detection methods and other representative methods~\cite{chen2023voxelnext,shi2021pv} on the Waymo \emph{validation} dataset~\cite{sun2020scalability} in terms of performance and running speed. 
It can be clearly observed that our {\ourMethod}-S~(\textit{i.e.}, the small version) not only surpasses existing DETR-based approaches in detection performance but also maintains a superior running speed.
In summary, our contributions are as follows:
\begin{itemize}

\item We introduce a novel dual query selection module, producing high-quality queries in a coarse-to-fine manner. 
\item We adopt an effective deformable grid attention module, which adaptively aggregates crucial regions and performs informative query interaction by properly leveraging the geometric information of point clouds. 
\item The proposed {\ourMethod} achieves state-of-the-art performance for 3D object detection on both the large-scale Waymo~\cite{sun2020scalability} and nuScenes~\cite{caesar2020nuscenes} datasets.
\end{itemize}

%% file: contents/2_related_work.tex
\section{Related Work}
\label{sec:related}

\noindent\textbf{2D Object Detection with DETR.}
DETR~\cite{carion2020end} is an end-to-end object detector that takes objects as queries and utilizes the transformer to interact queries with image features. Besides, DETR abandons many hand-crafted operations (\textit{e.g.}, Anchor, NMS) and utilizes Hungarian Matching to achieve the ground-truth assignment. The elegant architecture proposed by DETR brings a new insight into the research of object detection, and many works~\cite{zhu2020deformable, meng2021conditional,gao2021fast, wang2022anchor, li2022dn, liu2022dab, zhang2022dino} improve DETR from different perspectives. Deformable DETR~\cite{zhu2020deformable} introduces deformable attention into DETR and greatly improves the convergence speed of DETR. DN-DETR~\cite{li2022dn} proposes the denoising training strategy, which effectively reduces the learning difficulty of bipartite graph matching. DINO~\cite{zhang2022dino} utilizes contrastive learning in denoising training to achieve better performance.

\noindent\textbf{LiDAR-based 3D Object Detection.}
3D object detectors in point clouds can be categorized into point-based and voxel-based categories. For point-based, most methods~\cite{shi2019pointrcnn,yang2019std,yang20203dssd,qi2019deep,he2020structure,zhang2022not,yang2022dbq,chen2022sasa,he2022voxel,liu2022epnet++} directly utilize a PointNet-like backbone~\cite{qi2017pointnet,qi2017pointnet++} to extract point features, which can keep precise geometric structure information. However, these methods usually need to sample points to reduce computational costs, which may lose some important information in point clouds. For voxel-based, most methods~\cite{zhou2018voxelnet,yan2018second,lang2019pointpillars,deng2021voxel,shi2020pv,guan2022m3detr,yang2023pvt,liu2023flatformer,zhang2024hednet,yang2022graph,yang20213d,liu2020tanet,shi2020points} quantify point clouds into regular grids and utilize a 3D sparse convolution backbone to extract grid features (e.g. Voxel and Pillar) efficiently. 

\noindent\textbf{3D Object Detection with DETR.}
Due to the powerful feature representation of transformer, many works~\cite{bai2022transfusion,yan2023cross,zhu2023conquer,chen2023focalformer3d,wang2023uni3detr} have been explored to utilize DETR for 3D object detection in point clouds, especially for the design of two key components~(query selection and query interaction) in DETR.
Specifically, TransFusion~\cite{bai2022transfusion} selects the local maximum feature in BEV features as queries based on the heatmap. CMT~\cite{yan2023cross} adopts learnable queries initialized by 3D grids and utilizes global attention to interact queries with BEV features. ConQueR~\cite{zhu2023conquer} proposes a query contrast mechanism to reduce false positives. FocalFormer3D~\cite{chen2023focalformer3d} utilizes a multi-stage heatmap for better query selection. Besides, FocalFormer3D~\cite{chen2023focalformer3d} adopts deformable attention for efficient query interaction. Although the above DETR-based methods have made some progress, they are still inferior to some advanced methods~\cite{wang2023dsvt,chen2023voxelnext} that do not belong to the DETR-based paradigm. In this paper, we propose a simple and effective 3D DETR named {\ourMethod}, involving a novel dual query selection module for picking out high-quality queries and a deformable grid attention module to make effective query interaction by leveraging the rich geometric information of point clouds. 

%% file: contents/3_method.tex
\begin{figure*}[t!]
\centering
\includegraphics[width=0.99\linewidth]{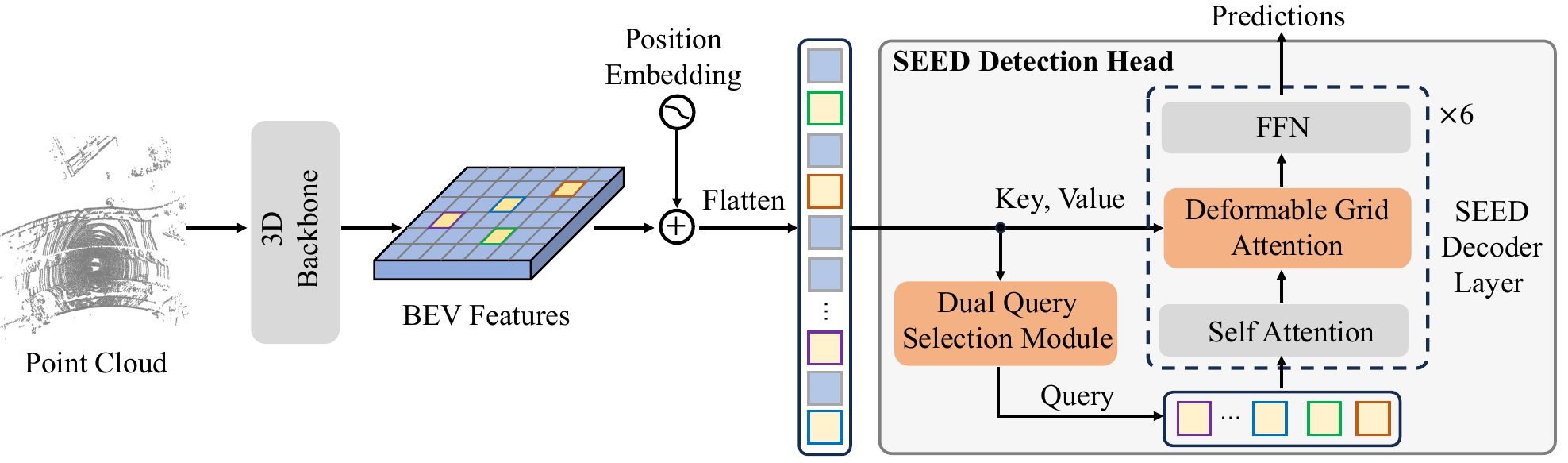}
\caption{Overall architecture of {\ourMethod}, which consists of a 3D backbone and a {\ourMethod} detection head. Specifically, the proposed {\ourMethod} detection head mainly includes a dual query selection~(DQS) module and a transformer decoder. The DQS utilizes a coarse-to-fine query selection strategy to select high-quality queries. The transformer decoder, including six {\ourMethod} decoder layers, takes these queries as inputs and then iteratively performs a self-attention operation for inter-query interaction and a proposed deformable grid attention~(DGA) for feature interaction between query and BEV features, generating final detection results.}
\label{fig_arch}
\vspace{-10pt}
\end{figure*}

\section{Method}
\label{sec:method}
Although many attempts have been made on DETR-based 3D object detection, there is still a certain performance gap with existing advanced LiDAR-based 3D detectors~\cite{chen2023voxelnext,wang2023dsvt}. We argue that the
main challenges come from two aspects. On the one hand, selecting superior queries from the high sparsity and uneven distribution of point clouds is not trivial. On the other hand, exploring how to make use of the rich geometric structure information from point clouds to perform effective query interaction is still challenging.

To mitigate these issues, we propose a \textbf{S}imple and \textbf{E}ff\textbf{E}ctive 3D \textbf{D}ETR method~(\textbf{\ourMethod}) for detecting 3D objects from point clouds. As shown in Figure~\ref{fig_arch}, we present the overall pipeline of {\ourMethod}. Specifically, we first feed point clouds into a classic voxel-based 3D backbone~\cite{yan2018second} to extract 3D voxel features and further convert them to BEV features. To retain their position information, we add position embedding to the BEV features. Then, the BEV features are flattened for subsequent query selection.
As for query selection, we propose a novel dual query selection~(DQS), which adopts a coarse-to-fine manner to obtain high-quality queries. Finally, the transformer decoder, including six {\ourMethod} decoder layers, is adopted to achieve feature interaction between the high-quality queries and the flattened BEV features, producing final detection results. In particular, our {\ourMethod} decoder layer leverages an effective deformable grid attention~(DGA) for query interaction instead of the cross attention operation in classic DETR decoder~\cite{carion2020end}. In the following, we will introduce the details of the proposed DQS and DGA in {\ourMethod}.



\subsection{Dual Query Selection Module}

\begin{figure}[h!]
\vspace{-10pt}
\centering
\includegraphics[width=0.8\linewidth]{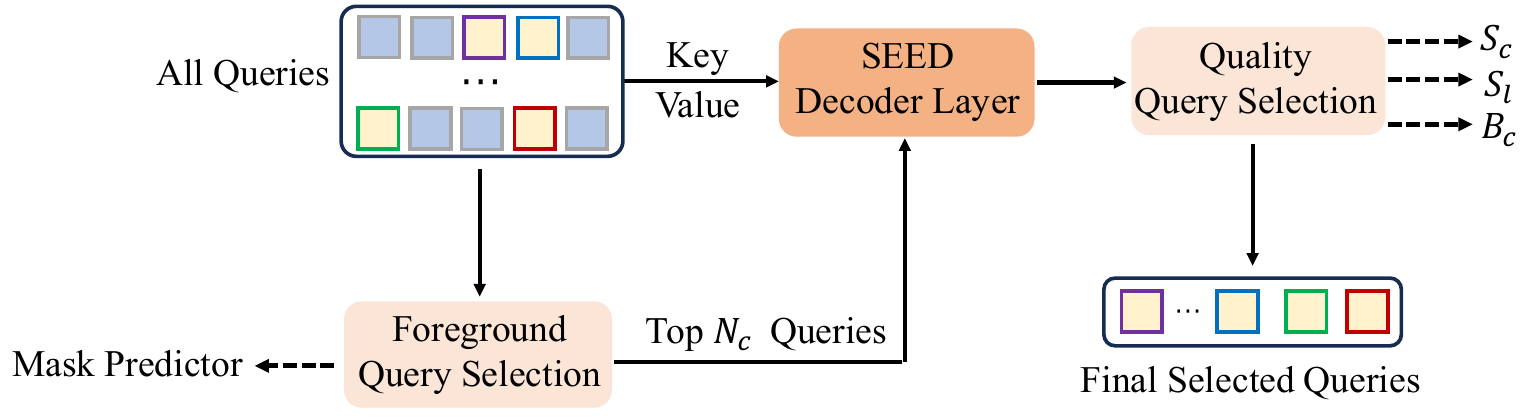}
\caption{Illustration of dual query selection~(DQS). DQS adopts a coarse-to-fine manner, which consists of a foreground query selection and a quality query selection. $S_c$, $S_l$, and $B_c$ are the predicted classification score, localization score, and regression for proposal boxes through three feed-forward networks~(FFN) branches, respectively. } 
\label{fig_dqs}
\vspace{-10pt}
\end{figure}

A proper query selection has demonstrated its importance in DETR-based 2D object detectors~\cite{liu2022dab,zhang2022dino,chen2023enhanced} to ensure accurate object localization and accelerate model convergence. However, due to the huge difference in data format between 2D images and 3D point clouds, it is necessary to consider some characteristics of point clouds, such as high sparsity and uneven distribution, for the query selection.  
Toward this goal, we propose a novel dual query selection~(DQS) module whose main purpose is to obtain high-quality queries in a coarse-to-fine manner. We present the detailed structure of DQS in Figure~\ref{fig_dqs}, which involves a foreground query selection and a quality query selection.

\noindent\textbf{Foreground Query Selection.} First, for the foreground query selection, we utilize a binary classification predictor to distinguish backgrounds and foregrounds on the BEV features. Simultaneously, we add BEV position embedding to BEV features and flatten them along the spatial dimension to generate all queries~(also named flattened BEV features). For the convenience of description, we define the flattened BEV features as $\bm{F}_{bev} \in \mathbb{R}^{(H \times W) \times C} $, where $H$, $W$ and $C$ are the height, width, and channel dimension of BEV features, respectively.
Then, we select these queries with proportion $r$ among the top confidence scores of BEV features $\bm{S}_{bev}$ from a mask predictor as the coarse queries, which can remain as many potential foreground queries as possible to ensure a high recall rate. Finally, the foreground query selection is formulated as:
\begin{equation}\label{q_coarse}
\bm{Q}_{c} = \mathrm{Top}_{N_{c}}(\bm{F}_{bev}, \bm{S}_{bev}),
\end{equation}
where $N_{c} = H \times W \times r$ and $\mathrm{Top}_{N_{c}}(x,y)$ means select top ${N_{c}}$ queries from $x$ according to $y$, and $N_{c}$ is the number of coarse queries.

After getting coarse queries, we further feed them to a {\ourMethod} decoder layer to achieve sufficient feature interaction between queries and flattened BEV features, producing the enhanced queries $\bm{Q}^{'}_{c}$, which can be computed as:
\begin{equation}\label{q_enchance}
\bm{Q}^{'}_{c} = \mathrm{Decoder}(\bm{Q}_{c}, \bm{F}_{bev}),
\end{equation}
where $\mathrm{Decoder}$ is our {\ourMethod} Decoder layer, which will be introduced in detail in the Section ~\ref{dga_method}.

\noindent\textbf{Quality Query Selection.} We first feed the coarse queries $\bm{Q}^{'}_{c}$ into three feed-forward networks~(FFN) branches to produce the classification score $\bm{S}_c$, the localization scores $\bm{S}_l$ and the regression $\bm{B}_c$ for coarse proposal boxes, whose corresponding ground truths are assigned based on the proposed quality-aware Hungarian Matching~(refer to Section~\ref{method:qhm}). Here, the classification score is the probability of recognizing 3D object proposals, and the localization score is defined as the 3D IoU of proposal boxes and the ground truths. Considering that the localization score is mainly for foreground objects, we set a proper classification score threshold $\tau$ to distinguish the foreground objects. Therefore, the quality scores $\bm{S}_q$ by combining these two indicators can be formulated as:
\begin{equation}
\bm{S}_{q}^{i}=\begin{cases}
(\bm{S}_c^{i})^{1-\beta} \cdot (\bm{S}_l^{i})^{\beta} \ \ ,  &  \mathrm{if} \ \bm{S}_c^{i}>\tau , \\
\bm{S}_c^{i} \ \ , & \text{otherwise},
\end{cases}
\end{equation}
where $\beta \in (0,1)$ is a hyper-parameter and is applied to control the importance of classification score and  localization score, and $i=0,1,...,N_c$. Then, we select the top $N_f$ fine proposal boxes $\bm{B}_{f}$ according to the quality scores $\bm{S}_q$ and concatenate them with the corresponding box quality scores $\bm{S}_{f}$. Next, we feed the concatenated features in a multi-layer perceptron~(MLP) to generate the geometric-aware high-quality queries. These steps can be formulated as:
\begin{equation}
\bm{B}_{f} = \mathrm{Top}_{N_f}(\bm{B}_c, \bm{S}_{q}),
\end{equation}
\begin{equation}
\bm{Q}_{f} = \operatorname{MLP}(\operatorname{Concat}(\bm{B}_f, \bm{S}_{f})).
\end{equation}
Finally, the output queries $\bm{Q}_{f}$ of DQS will be input to the subsequent SEED decoder.


\begin{figure}[t!]
\vspace{-10pt}
\centering
\includegraphics[width=0.99\linewidth]{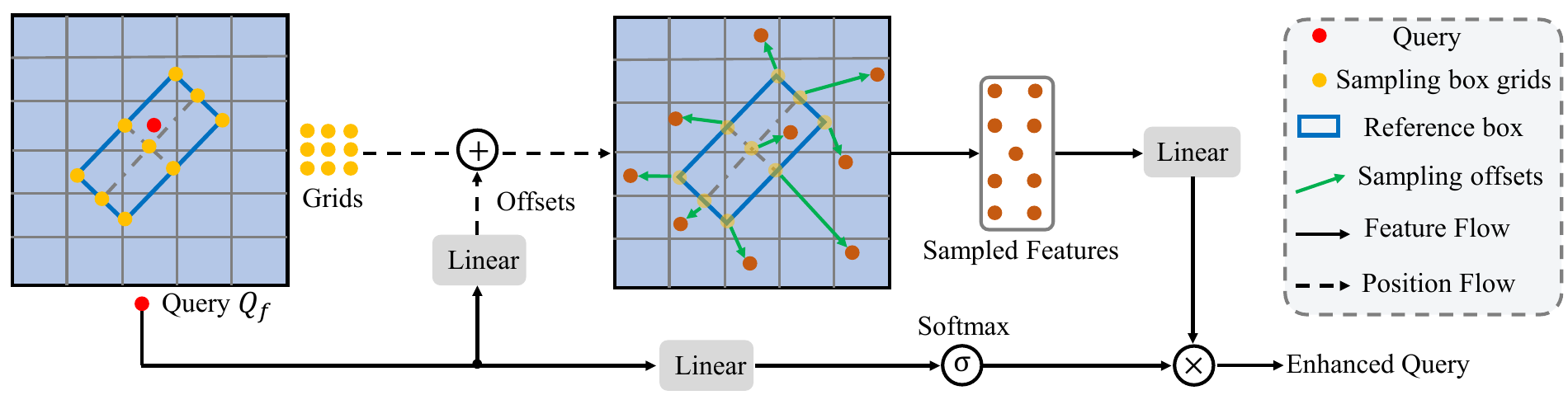}
\caption{Illustration of deformable grid attention~(DGA). DGA first uniformly divides each reference proposal into grids as the reference points and then utilizes the predicted offsets to achieve a flexible receptive field. 
}
\label{fig_dga}
\vspace{-10pt}
\end{figure}

\subsection{{\ourMethod} Decoder Layer} \label{dga_method}
The proposed {\ourMethod} decoder layer is applied to further enhance query feature representation by a self-attention operation and a cross attention operation and then maps the enhanced queries to task-specified outputs by an FFN. Different from existing DETR-based methods~\cite{yan2023cross,chen2023focalformer3d,zhu2023conquer}, we adopt a new cross attention operation in our {\ourMethod} decoder layer, namely deformable grid attention~(DGA). Next, we will introduce the reasons and the details of DGA.


\noindent\textbf{Why need DGA?}
An effective query interaction designed for point clouds is necessary to further explore the potential of the DETR paradigm in 3D detection. First of all, unlike 2D images, a nearby object may occupy most of the whole image, which even requires a global receptive field to detect the object well. However, a 3D object usually only occupies a small local area~(also be mentioned in SST~\cite{fan2022embracing}), which is much smaller than the range of the entire point clouds. Thus, the local attention may be enough for query interaction in point clouds. Second, point clouds possess rich geometric structures, especially for regular vehicles. Therefore, it is important to rationally utilize the geometric information of 3D objects. Third, although an accurate 3D proposal box can describe the geometric information of 3D objects, it is sub-optimal to capture some irregular objects or some hard objects. This indicates that a flexible receptive field is desired. Towards this goal, we propose a deformable grid attention~(DGA), which is a new local attention but adopts a flexible receptive field to effectively leverage the geometric information of 3D objects for query interaction.

\noindent\textbf{Details of DGA}. As shown in Figure~\ref{fig_dga}, we present the detailed structure of DGA. Specifically, we first regard the estimated proposal boxes $\bm{B}_f$ from DQS as the reference boxes and uniformly divide each reference box into $k \times k$ grids $\bm{g}_k$~(\textit{i.e.}, yellow points in Figure~\ref{fig_dga}). Then, we feed the corresponding selected queries $\bm{Q}_f$~(\textit{i.e.}, red points in Figure~\ref{fig_dga}) from DQS into a linear function, producing the predicted offsets $\Delta \bm{g}$. 
Next, we add the offsets to the grids $\bm{g}$ to generate the final sampling positions, which can capture the geometric information of 3D objects in a flexible receptive field. Meanwhile, the attention weight $\bm{A}$ is predicted by feeding $\bm{Q}_f$ into a linear function and a softmax function. Finally, the sampled features are multiplied with $\bm{A}$ to obtain the enhanced queries. We can formulate the above process of DGA as follows:
\begin{equation}\label{attn}
\mathrm{DGA}(\bm{g}, \bm{F}_{bev}) = \sum\limits_{j=1}^{K} \bm{A}_{j} \cdot \phi( \bm{F}_{bev}(\bm{g}_{j} + \Delta \bm{g}_{j})),
\end{equation}
where $K=k \times k$ and $\phi$ is a linear function for transforming sampled features to the attention space. $\bm{F}_{bev}(*)$ denotes sampling the corresponding features of the positions $*$ on the BEV features $\bm{F}_{bev}$ by a bilinear interpolation operation. Besides, we provide more comparisons with different attention operations~\cite{zhu2020deformable,nguyen2022boxer} in our supplemental materials.

\subsection{Quality-aware Hungarian Matching}\label{method:qhm}
Different from the traditional Hungarian Matching~\cite{carion2020end}, we introduce a quality-aware Hungarian Matching~(QHM) to effectively assign the ground truth. Specifically, QHM adopts the quality scores $\bm{S}_{f}$ instead of the classic classification scores in computing classification cost. Thus, our classification cost $\bm{\mathcal{C}}_{cls}$ can be formulated as: 
\begin{equation}
\bm{\mathcal{C}}_{pos} = -(1-\alpha)\cdot (\bm{S}_{f})^{\gamma} \cdot \log(1-\bm{S}_{f}),
\end{equation}
\begin{equation}
\bm{\mathcal{C}}_{neg} = -\alpha \cdot (1-\bm{S}_{f})^{\gamma} \cdot \log {\bm{S}_{f}},
\end{equation}
\begin{equation}
\bm{\mathcal{C}}_{cls} = \bm{\mathcal{C}}_{pos} - \bm{\mathcal{C}}_{neg},
\end{equation}
where $\alpha$, $\gamma$ are hyper-parameters. Finally, the total matching cost in Hungarian Matching can be computed as:
\begin{equation}
\bm{\mathcal{C}}_{\mathrm{match}}=\lambda_{cls}\cdot \bm{\mathcal{C}}_{cls}+\lambda_{reg}\cdot \bm{\mathcal{C}}_{reg}+\lambda_{giou}\cdot \bm{\mathcal{C}}_{giou},
\label{equation_cost}
\end{equation}
where $\bm{\mathcal{C}}_{reg}$ and $\bm{\mathcal{C}}_{giou}$ denote the regression cost and the GIoU cost, which have the same formulation in the traditional Hungarian Matching~\cite{carion2020end}. $\lambda_{cls}$, $\lambda_{reg}$, and $\lambda_{giou}$ are the balanced weights. 

%% file: contents/4_exps.tex
\section{Experiments}
\label{sec:exps}

\subsection{Datasets and Evaluation Metrics}
\noindent\textbf{Waymo Open Dataset.} 
Waymo Open Dataset~(WOD)~\cite{sun2020scalability} is a well-known and challenging large-scale outdoor 3D object detection benchmark. WOD consists of 1150 scenes (more than 200K frames), which are divided into three parts: 798 for training, 202 for validation, and 150 for testing. Each scene provides point clouds acquired by 64-beam LiDAR and covers a perception range with a size of 150m × 150m. Besides, the evaluation of WOD is divided into two levels~(LEVEL 1 and LEVEL 2) according to the number of points of the object. For all experimental results, we follow the standard protocol as the evaluation metric, which adopts 3D mean Average Precision~(mAP)~\cite{lin2014microsoft} and its weighted variant by heading accuracy~(mAPH) for three categories: \emph{Vehicle}, \emph{Pedestrian} and \emph{Cyclist}.

\noindent\textbf{nuScenes Dataset.}  nuScenes~\cite{caesar2020nuscenes} is another widely used autonomous driving dataset for LiDAR-based 3D object detection. The nuScenes dataset consists of 1000 scenes, which are divided into three parts: 750 for training, 150 for validation, and 150 for testing. Each scene is roughly 20s long, annotated at 2Hz, and provides point clouds collected by 32-beam LiDAR. Besides, the evaluation of the nuScnes dataset adopts the Mean Average Precision~(mAP) and nuScenes detection score~(NDS) to evaluate the performance of 3D detectors for 10 foreground classes.

\subsection{Implementation Details}
\noindent\textbf{Network Architecture.} In {\ourMethod}, we provide three versions, namely small~(\textit{i.e.}, {\ourMethod}-S), base~(\textit{i.e.}, {\ourMethod}-B), and large~(\textit{i.e.}, {\ourMethod}-L). In this paper, 
{\ourMethod}-S adopts the same 3D backbone as CenterPoint~\cite{yin2021center} for a fair comparison with most existing methods. For {\ourMethod}-B, we follow VoxelNext~\cite{chen2023voxelnext}, which introduces multi-scale voxel feature extraction and doubles the channel of the 3D backbone based on {\ourMethod}-S to improve the feature representation. For {\ourMethod}-L, we utilize a smaller voxel size (0.08, 0.08, 0.15) instead of (0.1, 0.1, 0.15) to enlarge the BEV resolution based on {\ourMethod}-B, which further boosts detection performance specifically for small objects such as \emph{Pedestrian}. 
In DQS, we set $r=0.3$ in foreground query selection to ensure a high recall and set $N_f=1000, \tau=0.2$ in quality query selection for effective query interaction in the subsequent transformer decoder. In DGA, we set $k = 5$ to divide grids as default.
In the transformer decoder, we adopt six {\ourMethod} decoder layers to iteratively perform query interaction and then generate the final 3D detection results. 

\noindent\textbf{Training.} The final loss includes the DETR-head loss~(similar to ConQueR~\cite{zhu2023conquer}) and our DQS loss. In DQS loss, we supervise  classification score with binary cross-entropy loss, localization score with IoU loss, and regression with Smooth-L1 loss, respectively. 
On WOD, we adopt the same point cloud range and data augmentations as CenterPoint~\cite{yin2021center}. Our model is optimized by AdamW optimizer~\cite{loshchilov2017decoupled} with the initial learning rate, weight decay, and momentum factor set to 0.001, 0.01, and 0.9, respectively. We train our model with a batch size of 24 on  8 NVIDIA Tesla V100 GPUs.  We run 24 epochs for the 20\% training set and only 12 epochs for the 100\% training set.
Besides, we utilize the fade strategy~\cite{wang2021pointaugmenting} to avoid over-fitting in the last epoch and the query contrast strategy~\cite{zhu2023conquer} to achieve better performance. For quality-aware Hungarian Matching, we set $\alpha$ to 0.25 and $\gamma$ to 2.0. In quality query selection, the hyper-parameter $\beta$ is set to 0.68, 0.71, and 0.65 for \emph{Vehicle}, \emph{Pedestrian}, and \emph{Cyclist}, respectively. This is a common setting in some non-DETR methods~(\textit{e.g.}, AFDetv2~\cite{hu2022afdetv2}, PillarNet~\cite{shi2022pillarnet}) for IoU-rectification. For matching cost, we set $\lambda_{cls}$, $\lambda_{reg}$, and $\lambda_{giou}$ to 1, 2, and 4. For nuScenes, our {\ourMethod} simply follows the settings of \cite{bai2022transfusion,chen2023focalformer3d}, including the range of point clouds, the voxel size, data augmentations, and the training strategy.

\begin{table*}[t!]
\caption{Performances on the Waymo Open Dataset \emph{validation} split (train with 100\% training data).  $\ddag$ denotes the two-stage method. `SEED-S', `SEED-B', and `SEED-L' mean the small, base, and large versions of {\ourMethod}, respectively. \textbf{Bold} denotes the best performance in DETR-based methods. All results are presented with single-frame input, no test-time augmentation, and no model ensembling.
}
\vspace{-10pt}
\footnotesize
\centering
\resizebox{1.0\linewidth}{!}{
\begin{tabular}{l|c|c|c|c|c|c|c|c|>{\columncolor[gray]{0.95}}c}
\specialrule{1pt}{1pt}{0pt}
\multicolumn{1}{c|}{\multirow{2}{*}{Methods}} & \multirow{2}{*}{Present at} & \multirow{2}{*}{DETR}  & \multicolumn{2}{c|}{\emph{Vehicle} 3D AP/APH} & \multicolumn{2}{c|}{\emph{Pedestrian} 3D AP/APH} & \multicolumn{2}{c|}{\emph{Cyclist} 3D AP/APH} & \multirow{2}{*}{\shortstack[1]{mAP/mAPH\\ L1}} \\
 & & & L1 & L2 & L1 & L2 & L1 & L2 & L2 \\
\midrule
SECOND~\cite{yan2018second} & Sensors 18 & \multirow{15}{*}{\XSolid}  & 72.3/71.7 & 63.9/63.3 & 68.7/58.2 & 60.7/51.3 & 60.6/59.3 & 58.3/57.0 & 61.0/57.2 \\
PointPillars~\cite{lang2019pointpillars} & CVPR 19 &  & 72.1/71.5 & 63.6/63.1 & 70.6/56.7 & 62.8/50.3 & 64.4/62.3 & 61.9/59.9 & 62.8/57.8\\ 
CenterPoint~\cite{yin2021center}& CVPR 21 &  & 74.2/73.6 & 66.2/65.7 & 76.6/70.5 & 68.8/63.2 & 72.3/71.1 & 69.7/68.5 & 68.2/65.8 \\
PV-RCNN$\ddag$~\cite{shi2020pv} & CVPR 20 &  & 78.0/77.5 & 69.4/69.0 & 79.2/73.0 & 70.4/64.7 & 71.5/70.3 & 69.0/67.8 & 69.6/67.2\\
SST\_TS$\ddag$~\cite{fan2022embracing} & CVPR 22 & & 76.2/75.8 & 68.0/67.6 & 81.4/74.0 & 72.8/65.9 & \textbf{--}/\textbf{--} & \textbf{--}/\textbf{--}  & \textbf{--}/\textbf{--} \\
AFDetV2~\cite{hu2022afdetv2} & AAAI 22 &  & 77.6/77.1 & 69.7/69.2 & 80.2/74.6 & 72.2/67.0 & 73.7/72.7 & 71.0/70.1 & 71.0/68.8 \\
SWFormer~\cite{Sun2022SWFormerSW} & ECCV 22 &  & 77.8/77.3 & 69.2/68.8 & 80.9/72.7 & 72.5/64.9 & \textbf{--}/\textbf{--} & \textbf{--}/\textbf{--} & \textbf{--}/\textbf{--} \\
PillarNet-34~\cite{shi2022pillarnet} & ECCV 22 &  & 79.1/78.6 & 70.9/70.5 & 80.6/74.0 & 72.3/66.2 & 72.3/71.2 & 69.7/68.7 & 77.3/74.6 \\
CenterFormer\cite{zhou2022centerformer} & ECCV 22 &  & 75.0/74.4 & 69.9/69.4 & 78.6/73.0 & 73.6/68.3 & 72.3/71.3 & 69.8/68.8 & 71.1/68.9\\ 
PV-RCNN++$\ddag$~\cite{shi2021pv} & IJCV 22 &  & 79.3/78.8 & 70.6/70.2 & 81.3/76.3 & 73.2/68.0 & 73.7/72.7 & 71.2/70.2 & 71.7/69.5\\ 
FSD$\ddag$~\cite{fan2022fully} & NeurIPS 22 &  & 79.2/78.8 & 70.5/70.1 & 82.6/77.3 & 73.9/69.1 & 77.1/76.0 & 74.4/73.3 & 72.9/70.8 \\
OcTr~\cite{zhou2023octr} & CVPR 23 &  & 78.1/77.6 & 69.8/69.3 & 80.8/74.4 & 72.5/66.5 & 72.6/71.5 & 69.9/68.9 & 70.7/68.2\\ 
PillarNeXt~\cite{li2023pillarnext} & CVPR 23 &  & 78.4/77.9 & 70.3/69.8 & 82.5/77.1 & 74.9/69.8 & 73.2/72.2 & 70.6/69.6 & 71.9/69.7\\ 
VoxelNext~\cite{chen2023voxelnext} & CVPR 23 &  & 78.2/77.7 & 69.9/69.4 & 81.5/76.3 & 73.5/68.6 & 76.1/74.9 & 73.3/72.2 & 72.2/70.1 \\ 
DSVT-Pillar~\cite{wang2023dsvt} & CVPR 23 &  & 79.3/78.8 & 70.9/70.5 & 82.8/77.0 & 75.2/69.8 & 76.4/75.4 & 73.6/72.7 & 73.2/71.0\\ 
DSVT-Voxel~\cite{wang2023dsvt} & CVPR 23 & & 79.7/79.3 & 71.4/71.0 & 83.7/78.9 & 76.1/71.5 & 77.5/76.5 & 74.6/73.7 & 74.0/72.1\\ 
\midrule
BoxeR-3D~\cite{nguyen2022boxer} & CVPR 22 & \multirow{7}{*}\Checkmark  & 70.4/70.0 & 63.9/63.7 & 64.7/53.5 & 61.5/53.7 & 50.2/48.9 & \textbf{--}/\textbf{--} & \textbf{--}/\textbf{--} \\
TransFusion~\cite{bai2022transfusion} & CVPR 22 &  & \textbf{--}/\textbf{--} & \textbf{--}/65.1 & \textbf{--}/\textbf{--} & \textbf{--}/63.7 & \textbf{--}/\textbf{--} & \textbf{--}/65.9 & \textbf{--}/64.9 \\ 
ConQueR~\cite{zhu2023conquer} & CVPR 23 &  & 76.1/75.6 & 68.7/68.2 & 79.0/72.3 & 70.9/64.7 & 73.9/72.5 & 71.4/70.1 & 70.3/67.7 \\
FocalFormer3D~\cite{chen2023focalformer3d} & ICCV 23 &  & \textbf{--}/\textbf{--} & 68.1/67.6 & -/- & 72.7/66.8 & \textbf{--}/\textbf{--} & 73.7/72.6 & 71.5/69.0\\ 
{\ourMethod}-S~(Ours) & \textbf{--} &  & 78.2/77.7 & 70.2/69.7 & 81.3/75.8 & 73.3/68.1 & 78.4/77.2 & 75.7/74.5 & 73.1/70.8 \\
{\ourMethod}-B~(Ours) & \textbf{--} & & 79.7/79.2 & 71.8/71.4 & 83.1/78.3 & 75.5/70.8 & 80.0/78.8 & 77.3/76.1 & 74.9/72.8\\ 
{\ourMethod}-L~(Ours) & \textbf{--} &  & \textbf{79.8/79.3} & \textbf{71.9/71.5} & \textbf{83.6/79.1} & \textbf{76.2/71.8} & \textbf{81.2/80.0} & \textbf{78.4/77.3} & \textbf{75.5/73.5} \\ 
\bottomrule
\end{tabular}
}
\vspace{-5pt}
\label{tab:sota}
\end{table*}

\begin{table}[t]
    \centering
    \caption{Effectiveness of our {\ourMethod} with multiple frames as inputs on the Waymo Open Dataset \emph{validation} and \emph{test} split.}
    \vspace{-10pt}
    \subfloat[Effectiveness of {\ourMethod} on \emph{validation} split.]{
    \resizebox{0.47\linewidth}{!}{
     \begin{tabular}{l|c|c|c}
        \toprule
         Methods  & Frames &mAP/mAPH~(L1)  &mAP/mAPH~(L2)  \\
        \midrule
        CenterPoint~\cite{yin2021center} & 4 & 76.4/74.9 & 70.8/69.4\\
        CenterFormer~\cite{zhou2022centerformer} & 4 & 78.5/77.0 & 74.7/73.2 \\
        MPPNet~\cite{chen2022mppnet} & 4 & 81.1/79.9 & 75.4/74.2\\
        MSF~\cite{he2023msf} & 4 & 81.1/80.2 & 76.0/74.6 \\
        PillarNeXt~\cite{shi2022pillarnet} & 3 & 81.5/80.0 & 75.9/74.5   \\
        DSVT-Voxel~\cite{wang2023dsvt}  & 3  &  82.1/80.8 & 76.3/75.0    \\
        \hline
        SEED-S~(Ours)  & 3  &  81.6/80.1 &  75.8/74.3     \\
        SEED-B~(Ours)  & 3  &  82.9/81.4 &  77.2/75.8     \\
        SEED-L~(Ours)  & 3  &  \textbf{83.1/81.6} &  \textbf{77.5/76.1}     \\
        \bottomrule
    \end{tabular}
    }
    \label{tab:mf_val}
    }
    \subfloat[Effectiveness of {\ourMethod} on \emph{test} benchmark.]{
    \resizebox{0.48\linewidth}{!}{
    \begin{tabular}{l|c|c|c}
        \toprule
         Methods  & Frames &mAP/mAPH~(L1)  &mAP/mAPH~(L2)  \\
        \midrule
        PV-RCNN++~\cite{shi2021pv} & 1 & 78.0/75.7 & 72.4/70.2 \\
        AFDetV2~\cite{hu2022afdetv2} & 1 & 77.6/75.2 & 72.2/70.3\\
        PillarNet~\cite{shi2022pillarnet} & 1 & 77.5/74.7 & 72.2/69.6\\
        FSD~\cite{fan2022fully} & 1 & 80.4/78.2 & 74.4/72.4\\
        ConQueR~\cite{zhu2023conquer} & 1 & \textbf{--}/\textbf{--} & \textbf{--}/72.0 \\
        SEED-L~(Ours)  & 1  & \textbf{81.7/79.7} & \textbf{76.5/74.5}     \\
        \hline
        CenterPoint++~\cite{yin2021center} & 3 & 79.4/77.9 & 74.2/72.8\\
        PillarNeXt~\cite{li2023pillarnext} & 3 & 80.5/79.0 & 75.5/74.1 \\
        SEED-L~(Ours)  & 3  &  \textbf{83.5/82.1} &  \textbf{78.7/77.3}     \\
        \bottomrule
        \end{tabular}
    }
    \label{tab:test}
    }
\vspace{-15pt}
\end{table}

\subsection{Main Results}
\noindent\textbf{Results on WOD.} 
We present the comparison with existing DETR-based methods~(bottom) and other representative methods~(top) on the WOD in Table~\ref{tab:sota}. Here, we provide three versions of {\ourMethod}, including the small~({\ourMethod}-S), base~({\ourMethod}-B) and large~({\ourMethod}-L).
Compared with DETR-based 3D detectors, our {\ourMethod}-S outperforms the advanced ConQueR~\cite{zhu2023conquer} and FocalFormer3D~\cite{chen2023focalformer3d} with 3.1 and 1.8 mAPH/L2, respectively. Moreover, {\ourMethod}-S possesses a satisfactory running speed with about 13.5 FPS on an NVIDIA GeForce RTX 3090~(see Figure~\ref{fig_intro}). These benefits on both the detection performance and the running speed effectively illustrate the superiority of {\ourMethod}.
Besides, we compare our {\ourMethod} with other advanced methods that do not belong to the DETR paradigm. {\ourMethod}-B exceeds the representative two-stage methods PV-RCNN++~\cite{shi2021pv} with 3.3 mAPH/L2. Considering that the high-resolution feature map is necessary to detect small 3D objects, {\ourMethod}-L further boosts the detection performance of {\ourMethod}-B) with 1.0~(71.8 \textit{vs.} 70.8) APH/L2 on \emph{Pedestrian}. It is noteworthy that {\ourMethod}-L even outperforms the previous state-of-the-art~(SOTA) method DSVT-Voxel~\cite{wang2023dsvt} 
with 1.4~(73.5 \textit{vs.} 72.1) mAPH/L2, leading to a new SOTA. Note that the BEV resolution of {\ourMethod}-L is still much smaller than DSVT-Voxel~\cite{wang2023dsvt} since DSVT uses one-stride 3D backbone like SST~\cite{fan2022embracing} to improve the detection performance. Furthermore, DSVT~\cite{wang2023dsvt} focuses on enhancing the representation ability of the 3D backbone, which is orthogonal to our {\ourMethod}. 

In Table~\ref{tab:mf_val}, we also provide the results of our {\ourMethod} with three frames as inputs on the WOD \emph{validation} split. We observe that {\ourMethod}-L produces a leading performance with 76.1 mAPH/L2, even surpassing the temporal 3D object detection method MPPNet~\cite{chen2022mppnet} with an obvious margin.
To further verify the effectiveness of our {\ourMethod}, we evaluate the performance of our {\ourMethod}-L with one frame and three frames as inputs on the WOD \emph{test} benchmark, as shown in Table~\ref{tab:test}. {\ourMethod}-L with single frame achieves 74.5 mAPH/L2, which exceeds DETR-based method ConQueR~\cite{zhu2023conquer} with 2.5~(74.5 \textit{vs.} 72.0) mAPH/L2. For three frames, {\ourMethod}-L outperforms the representative method PillarNeXt~\cite{li2023pillarnext} with 3.2~(77.3 \textit{vs.} 74.1) mAPH/L2. All the experimental results clearly demonstrate the superiority of our {\ourMethod}.

\noindent\textbf{Results on nuScenes.} We also evaluate our {\ourMethod} on the \textit{validation} split of nuScenes dataset~\cite{caesar2020nuscenes} to further verify the effectiveness of our {\ourMethod}. As shown in Table~\ref{tab:nusc}, {\ourMethod} achieves 71.2 NDS and 66.6 mAP, which exceeds the popular DETR-based detector TransFusion-L~\cite{bai2022transfusion} with 1.1 NDS and 1.5 mAP under the same 3D backbone. This demonstrates the generalization of our method.

\begin{table*}[t]
\centering
\caption{
Comparison with state-of-the-art methods on the nuScenes \textit{validation} split. All results are presented without any test-time augmentation or model ensembling. Here, our {\ourMethod} adopts the same 3D backbone with CenterPoint~\cite{yin2021center} for a fair comparison. * denotes the reproduced result from official code.}
\vspace{-10pt}
\setlength{\tabcolsep}{10pt}
\resizebox{1.0\linewidth}{!}{
\begin{tabular}{c|c|ccccc|>{\columncolor[gray]{0.95}}c>{\columncolor[gray]{0.95}}c}
  \toprule
   Method & Present at & mATE & mASE & mAOE & mAVE & mAAE & NDS & mAP\\
    \midrule
     PointPillar~\cite{lang2019pointpillars} & CVPR 19 &  0.424 &0.284 &0.529 &0.377 &0.194 & 49.1 & 34.3\\
     CenterPoint~\cite{yin2021center} & CVPR 21 & 0.291 &0.252 &0.324 &0.284 &0.189 & 64.9 & 56.6\\
     TransFusion-L~\cite{bai2022transfusion} & CVPR 22 & \textbf{--} &\textbf{--} &\textbf{--} &\textbf{--} &\textbf{--}  & 70.1& 65.1\\
     PillarNet~\cite{shi2022pillarnet} & ECCV 22 & 0.277 &0.252 &0.289 &0.247 &0.191 & 67.4 & 59.8\\
     UVTR-L~\cite{li2022unifying} & NeurIPS 22 &  0.334 &0.257 &0.300 &0.204 &0.182 & 67.7 & 60.9\\
     VoxelNeXt*~\cite{chen2023voxelnext} & CVPR 23 &  0.301 &0.252 &0.406 &0.217 &0.186 & 66.7 & 60.5\\
     Uni3DETR~\cite{wang2023uni3detr} & NeurIPS 23 & 0.288 &0.249 &0.303 &0.216 &0.181 & 68.5 & 61.7\\
    {\ourMethod}~(Ours) & \textbf{--} & 0.279 & 0.257 & 0.284 & 0.208 & 0.187 & \textbf{71.2} & \textbf{66.6} \\
\bottomrule
\end{tabular}
}
\vspace{-10pt}
\label{tab:nusc}
\end{table*}


\subsection{Ablation Study}
In this section, we conduct extensive ablation studies to investigate the effectiveness of {\ourMethod} on the Waymo \textit{validation} set with 20\% training data by default if not specified. And we adopt {\ourMethod}-S as our default {\ourMethod} model in the following ablation studies. For more ablation studies about our {\ourMethod}, please refer to our supplemental materials.

\noindent\textbf{Effectiveness of the Proposed Components.} As shown in Table~\ref{tab:ab_com}, we conduct ablation studies on our proposed two components in {\ourMethod}.  
First, we set the baseline by replacing our DQS with heatmap-based query selection~\cite{bai2022transfusion} and replacing our DGA with box attention~\cite{nguyen2022boxer} in our SEED. 
Compared with the baseline, the proposed DQS module brings 2.8 mAPH/L2~(67.4 \textit{vs.} 64.6) performance improvement in average for \emph{Vehicle},  \emph{Pedestrian}, and  \emph{Cyclist} on Waymo dataset, which demonstrates the superiority of DQS for selecting out high-quality queries. Besides, the proposed DGA boosts the performance of the baseline with 1.8 mAPH/L2, demonstrating the effectiveness of DGA in achieving feature interaction. Finally, thanks to the benefits of the proposed DQS and DGA, {\ourMethod} has an obvious gain of 3.6 mAPH/L2 over the baseline model. 

\begin{table}[t]
\caption{Ablation study for each component in {\ourMethod}. We use mAP/mAPH~(L2) to evaluate the overall detection performance.}
\vspace{-10pt}
\small
\centering
\setlength{\tabcolsep}{20pt}
\resizebox{1.0\linewidth}{!}{
\begin{tabular}{l|c|c|c|c|c}
\toprule
\multicolumn{1}{c|}{\multirow{2}{*}{DQS}} & \multirow{2}{*}{DGA} & \multicolumn{3}{c|}{3D AP/APH~(L2)} & \multirow{2}{*}{\shortstack{mAP/mAPH \\ (L2)}}\\
 & & \emph{Vehicle} & \emph{Pedestrian} & \emph{Cyclist}\\
\midrule
\textbf{--}  & \textbf{--}  & 65.4/64.9 & 68.8/63.3  & 66.7/65.5 & 67.0/64.6  \\
$\checkmark$  & \textbf{--}  & 68.0/67.5 & 70.9/65.2 & 70.7/69.5 &69.9/67.4 \\
\textbf{--}  & $\checkmark$ & 65.7/65.2 & 69.9/64.4  & 71.0/69.6 &68.8/66.4  \\
$\checkmark$  & $\checkmark$  &\textbf{68.5/68.1} & \textbf{72.1/66.5} &\textbf{71.2/70.0}  &\textbf{70.6/68.2}  \\
\bottomrule
\end{tabular}
 }
\label{tab:ab_com}
\end{table}

\noindent\textbf{Superiority of the DQS.} 
To further demonstrate the superiority of the proposed DQS module, we provide three representative
query selection strategies for comparison with DQS in Table~\ref{tab:ab_dqs}, namely \textit{Learnable}, \textit{Heatmap-based} and \textit{Top-N} manners. Specifically, \textit{Learnable} means that we obtain queries by adopting a learnable manner like CMT~\cite{yan2023cross}. \textit{Heatmap-based}~(\textit{e.g.}, TransFusion-L~\cite{bai2022transfusion}) is a classic query selection strategy in LiDAR-based 3D object detection, which collects the local maximum elements in BEV features as queries. The advanced query selection manner \textit{Top-N}~(\textit{e.g.}, ConQueR~\cite{zhu2023conquer}) uses a class-agnostic feed-forward network~(FFN) head to obtain Top-N scored box proposals, which are selected as object queries. 
As shown in Table~\ref{tab:ab_dqs}, \textit{Heatmap-based} produces the worst performance among these methods. We think the main reason is that the queries~(as Query) of \textit{Heatmap-based} method is obtained directly from BEV features~(as Key, Value), rather than learnable queries or geometric-aware queries. This often leads to sub-optimal feature interaction, which makes it difficult for the subsequent decoder to stack more layers due to the potential risk of over-fitting.
However, our DQS respectively outperforms the \textit{Learnable} and \textit{Top-N} manners with 1.6 and 1.4 mAPH/L2, which illustrates the superiority of dual query selection for picking out high-quality geometric-aware queries.

\begin{table}[t]
\caption{Ablation study for DQS in {\ourMethod}. We adopt mAP/mAPH~(L2) to evaluate the detection performance.}
\vspace{-10pt}
\small
\centering
\setlength{\tabcolsep}{20pt}
\resizebox{1.0\linewidth}{!}{
\begin{tabular}{l|c|c|c|c}
\toprule
\multicolumn{1}{l|}{\multirow{2}{*}{Methods}} & \multicolumn{3}{c|}{3D AP/APH~(L2)} & \multirow{2}{*}{\shortstack{mAP/mAPH \\ (L2)}}\\
& \emph{Vehicle} & \emph{Pedestrian} & \emph{Cyclist}\\
\midrule
Learnable~\cite{yan2023cross}     &  66.1/65.6 & 70.1/64.5  & 71.0/69.8 & 69.1/66.6  \\
Heatmap-based~\cite{bai2022transfusion}   & 64.3/63.8 & 69.7/64.1  & 68.4/67.1 & 67.5/65.0 \\
Top-N~\cite{zhu2023conquer}   & 66.1/65.7 & 70.6/65.3  & 70.7/69.5  & 69.1/66.8 \\
DQS~(Ours)  &  \textbf{68.5/68.1} & \textbf{72.1/66.5} &\textbf{71.2/70.0}  &\textbf{70.6/68.2} \\
\bottomrule
\end{tabular}
 }
\label{tab:ab_dqs}
\vspace{-10pt}
\end{table}

\begin{table}[t]
\caption{Ablation study for DGA in {\ourMethod}. We adopt mAP/mAPH~(L2) to evaluate the detection performance.}
\vspace{-10pt}
\small
\centering
\setlength{\tabcolsep}{20pt}
\resizebox{1.0\linewidth}{!}{
\begin{tabular}{l|c|c|c|c}
\toprule
\multicolumn{1}{l|}{\multirow{2}{*}{Methods}} & \multicolumn{3}{c|}{3D AP/APH~(L2)} & \multirow{2}{*}{\shortstack{mAP/mAPH \\ (L2)}}\\
& \emph{Vehicle} & \emph{Pedestrian} & \emph{Cyclist}\\
 \midrule
Global Attention~\cite{vaswani2017attention}    & \textbf{--}~(OOM) & \textbf{--}~(OOM) & \textbf{--}~(OOM) & \textbf{--}~(OOM) \\
Deformable Attention~\cite{zhu2020deformable}    &  67.5/67.0 & 71.3/65.9  & 70.8/69.7 & 69.9/67.5     \\
Box Attention~\cite{nguyen2022boxer}   & 67.9/67.4 & 71.1/65.5 & 70.9/69.7 & 70.0/67.5 \\ 
DGA~(Ours)   &\textbf{68.5/68.1} & \textbf{72.1/66.5} &\textbf{71.2/70.0}  &\textbf{70.6/68.2}  \\
\bottomrule
\end{tabular}
 }
\label{tab:ab_dga}
\end{table}

\noindent\textbf{Effectiveness of the DGA.} To illustrate the effectiveness of the proposed DGA, we compare it with three representative query interaction operations, including global attention~\cite{vaswani2017attention}, deformable attention~\cite{zhu2020deformable} and box attention~(or named non-deformable grid attention)\cite{nguyen2022boxer}, whose results are summarized in Table~\ref{tab:ab_dga}. Since most LiDAR-based 3D detection methods~\cite{yin2021center,bai2022transfusion} have larger feature maps~(\textit{e.g.}, $180\times180$), performing global attention operation on mainstream GPUs~(\textit{e.g.}, NVIDIA Tesla V100 or NVIDIA GeForce RTX 3090) is unable to bear such a large computational cost, causing the GPU to run out of memory~(OOM). Therefore, the local attention operation, including the deformable attention and the box attention, is still reasonable to promise performance and efficiency for query interaction. For deformable attention operation, it produces 67.0 APH/L2 on \emph{Vehicle}, which is inferior to the box attention manner with 67.4 APH/L2. This indicates that effectively leveraging the box geometric information is important in feature interaction. However, the box attention operation depends on the precision of box regression, and its receptive field is not as flexible as deformable attention. Therefore, on hard objects such as \emph{Pedestrian}, the deformable attention outperforms the detection performance of the box attention. In contrast, our DGA has the advantages of both the flexible receptive field of deformable attention and the rich geometric information of box attention. Not surprisingly, our DGA surpasses these local attention methods, demonstrating its effectiveness.

\begin{table}[t]
\caption{Ablation study for quality-aware Hungarian Matching~(QHM) in {\ourMethod}. THM is short for traditional Hungarian Matching. Besides, we adopt mAP/mAPH~(L2) to evaluate the detection performance.}
\vspace{-10pt}
\small
\centering
\setlength{\tabcolsep}{20pt}
\resizebox{1.0\linewidth}{!}{
\begin{tabular}{l|c|c|c|c}
\toprule
\multicolumn{1}{l|}{\multirow{2}{*}{Methods}} & \multicolumn{3}{c|}{3D AP/APH~(L2)} & \multirow{2}{*}{\shortstack{mAP/mAPH \\ (L2)}}\\
& \emph{Vehicle} & \emph{Pedestrian} & \emph{Cyclist}\\
\midrule
THM\cite{carion2020end}    & 67.3/66.8 & 71.7/66.4  & 70.9/69.7 & 70.0/67.6\\
QHM~(Ours)   &\textbf{68.5/68.1} & \textbf{72.1/66.5} &\textbf{71.2/70.0}  &\textbf{70.6/68.2}  \\
\bottomrule
\end{tabular}
 }
\label{tab:hungarian}
\vspace{-10pt}
\end{table}

\noindent\textbf{Influence of Quality-aware Hungarian Matching.} Hungarian Matching is an indispensable component in the existing DETR-based approaches.  
In {\ourMethod}, we adopt a new matching manner, named quality-aware Hungarian Matching~(QHM). To verify the influence of QHM on 3D detection performance, we compare it with the traditional Hungarian Matching~\cite{carion2020end}~(THM), whose result is presented in Table~\ref{tab:hungarian}. It can be observed that QHM produces a consistent performance improvement over THM in terms of mAPH/L2, which benefits from taking the quality scores of 3D objects into account when computing classification cost in Hungarian Matching. Moreover, we carefully find there is a better gain on \emph{Vehicle} than \emph{Pedestrian} and \emph{Cyclist}. The main reason is that the localization score of \emph{Vehicle} is more easily estimated than that of \emph{Pedestrian} and \emph{Cyclist} due to some factors such as a large size and a rigid object. 

\begin{figure}[t!]
\centering
	\includegraphics[width=0.90\linewidth]{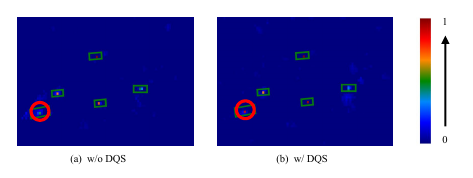}
	\vspace{-15pt}
	\caption{\small{
		Visualization of {\ourMethod} without DQS (the first row) and with DQS (the second row). We highlight the challenging queries with red circles. The colormap indicates the values of the confidence scores for selected queries on the BEV map. Green boxes are the ground truth boxes.
		}}\label{fig_vis}
	\vspace{-10pt}
\end{figure}

\subsection{Visualization Analysis}


As shown in Figure~\ref{fig_vis}, we present the visualization of our {\ourMethod} with and without DQS~(\textit{i.e.}, directly select Top $N_f$ queries in one step). Specifically, we visualize the positions of final selected queries from query selection on the BEV map, whose corresponding colormap denotes the confidence score of the selected queries. It can be observed that after utilizing DQS, some hard queries are successfully captured, which indicates that DQS can enhance the confidence score of some potential hard objects.
This phenomenon further demonstrates the superiority of our DQS. Besides, we provide more visualization of our {\ourMethod} in the supplemental materials.


\subsection{Limitation}

Our method mainly improves the detection head based on the DETR paradigm for 3D object detection. Therefore, the advanced 3D detectors that focus on enhancing the representation ability of 3D backbone are orthogonal to {\ourMethod}. In the future, we plan to apply our {\ourMethod} to more powerful 3D backbones on more datasets to further explore the scalability of our method. Besides, we observe that {\ourMethod} may fail to detect some distant and small 3D objects, but they are clearly visible in 2D camera images. Therefore, exploiting the complementarity of multiple modalities~(\textit{i.e.}, 3D point clouds, and 2D camera images) to detect these challenging objects is also our next step.

%% file: contents/5_conclusion.tex
\section{Conclusion}
\label{sec:conclusion}

In this paper, we have presented a simple and effective 3D DETR framework named {\ourMethod} to detect 3D objects from point clouds. Specifically, {\ourMethod} involves two key components:
a dual query selection~(DQS) module to retain high-quality queries in a coarse-to-fine manner and a deformable grid attention~(DGA) module to capture informative features by performing sufficient query interaction. Extensive ablation studies have demonstrated the effectiveness of the proposed DQS and DGA. Thanks to the superiority of the proposed DQS and DGA, our {\ourMethod} has achieved state-of-the-art 3D detection performance on the large-scale Waymo and nuScenes dataset. 
Finally, we hope {\ourMethod} could become a new strong baseline for the  community of DETR-based 3D object detection.

%

%% file: contents/6_supp.tex


\appendix
\section{Appendix}

The supplementary materials are organized as follows.
First, in section~\ref{sec:extra_exp}, we present the extra experiments of illustrating the capability of our {\ourMethod}, applying different backbones, varying the score threshold $\tau$ for quality query selection in DQS and the distinct grids in DGA on the Waymo validation set~\cite{sun2020scalability} with 20\% training data, respectively. Besides, we explore the impact of different numbers of {\ourMethod} decoder layers on detection performance.
In section~\ref{sec:diff_attn_exp}, we present the comparisons of several variant attention operations for query interaction. In section~\ref{sec:discussion}, we discuss the differences of our proposed  DQS and DGA with the existing related methods. 
Finally, we provide the analysis of visualization, including the learned attention map of DGA and the 3D detection results under different settings in section~\ref{sec:vis}.



\subsection{Extra Experiments}\label{sec:extra_exp}


\begin{table}[h!]
\caption{Effectiveness of our {\ourMethod}. For a fair comparison, we adopt 100\% Waymo training data for all models. The results are evaluated by the metric of mAP/mAPH~(L2).}
\vspace{-10pt}
\small
\centering
\setlength{\tabcolsep}{10pt}
\resizebox{1.0\linewidth}{!}{
\begin{tabular}{l|c|c|c|c|c}
\toprule
 Methods & Detection Head & mAP/mAPH~(L2) & FLOPs~(G) & Params~(M) & Latency~(ms) \\
\midrule
SECOND~\cite{yan2018second} & Anchor-based & 61.0/57.2 & 91.2 & 5.3 & 33.3  \\
CenterPoint~\cite{yin2021center}& Center-based & 68.2/65.8 & 141.2 & 7.8 & 44.0   \\
PV-RCNN++~\cite{shi2021pv} & RoI-based & 71.7/69.5 & 166.6 & 16.1 & 149.0  \\
VoxelNeXt~\cite{chen2023voxelnext} & Center-based & 72.2/70.1 & 624.9 & 29.3 & 124.7 \\
\midrule
TransFusion~\cite{bai2022transfusion} & DETR-based & \textbf{--}/64.9 & 96.8 & 7.9 & 70.5  \\
ConQueR~\cite{zhu2023conquer} & DETR-based & 70.3/67.7 & 167.3 & 15.1 & 99.1 \\
FocalFormer3D~\cite{chen2023focalformer3d} & DETR-based & 71.5/69.0 & 144.9 & 19.4 & 97.2 \\
\midrule
\textbf{{\ourMethod}-S~(Ours)} & DETR-based & \textbf{73.1/70.8} & 168.7 & 12.8 & 74.2 \\
\textbf{{\ourMethod}-L~(Ours)}  & DETR-based & \textbf{75.5/73.5} & 648.1 & 33.1 & 163.8 \\
\bottomrule
\end{tabular}
 }
\label{tab:ab_gen}
\vspace{-5pt}
\end{table}

\noindent\textbf{Capability of our {\ourMethod}.} To verify the capability of our {\ourMethod}, we adopt the small version  {\ourMethod}-S with the same 3D backbone as CenterPoint~\cite{yin2021center} for a fair comparison with existing representative 3D object detection methods, including anchor-based~\cite{yan2018second}, center-based~\cite{yin2021center}, RoI-based~\cite{shi2021pv} and DETR-based~\cite{bai2022transfusion,zhu2023conquer,chen2023focalformer3d} detectors. We conduct the comparisons of these methods in terms of performance, FLOPs, parameters, and latency, shown in Table~\ref{tab:ab_gen}.
Note that the main difference between these methods is the design of the detection head. Moreover, we evaluate the running speed of all approaches on one NVIDIA GeForce RTX 3090 with a batch size of 1 according to their corresponding official open-source code for a fair comparison. 
Compared with SECOND~\cite{yan2018second} and CenterPoint~\cite{yin2021center}, our {\ourMethod}-S has a slower running speed, but our performance greatly exceeds them with 13.6 and 5.0 mAPH/L2, respectively. Furthermore, benefiting from the well-designed DQS module for selecting high-quality queries and the superior DGA operation for effective feature interaction, the detection performance of our {\ourMethod}-S even outperforms PV-RCNN++~\cite{shi2021pv} of 1.3 mAPH/L2 with $2\times$ faster running speed. However, existing DETR-based methods still fall behind PV-RCNN++ in terms of detection performance. The above experimental results effectively illustrate the powerful capability of our {\ourMethod}.

\begin{table}[t!]
\caption{Effectiveness of our {\ourMethod} on different backbones on the Waymo validation set~\cite{sun2020scalability} with 20\% training data. We use mAP/mAPH~(L2) for evaluating the detection performance. $*$ means our reproduced performance from the official code.}
\vspace{-10pt}
\small
\centering
\setlength{\tabcolsep}{12pt}
\resizebox{1.0\linewidth}{!}{
\begin{tabular}{l|c|c|c|c}
\toprule
\multicolumn{1}{c|}{\multirow{2}{*}{Methods}} & \multicolumn{3}{c|}{3D AP/APH~(L2)} & \multirow{2}{*}{\shortstack{mAP/mAPH \\ (L2)}}\\
& \emph{Vehicle} & \emph{Pedestrian} & \emph{Cyclist}\\
\midrule
CenterPoint-Pillar~\cite{yin2021center}    &  62.2/61.7 & 65.1/55.0  & 63.0/61.5  & 63.4/59.4     \\
\textbf{+SEED Detection Head}    &67.0/66.5 & 71.3/62.0 &65.8/64.5  &\textbf{68.0/64.3}    \\
\hline
CenterPoint~\cite{yin2021center}   &  63.2/62.7 & 64.3/58.2  & 66.1/64.9 & 64.5/61.9     \\
\textbf{+SEED Detection Head}  &68.5/68.1 & 72.1/66.5 &71.2/70.0  &\textbf{70.6/68.2}   \\
\hline
DSVT-Pillar$*$~\cite{wang2023dsvt}  &  69.7/69.2 & 74.9/68.0  & 70.7/69.6  &71.8/68.9     \\
\textbf{+SEED Detection Head}   &  71.7/71.3 & 75.4/68.7  & 73.0/71.8  &\textbf{73.4/70.6}     \\
\hline
HEDNet$*$~\cite{zhang2024hednet}  &  70.8/70.3 & 75.0/70.3  & 73.6/72.6  &73.1/71.1     \\
\textbf{+SEED Detection Head}   &  72.4/72.0 & 76.3/71.3  & 74.9/73.8  &\textbf{74.5/72.4}     \\
\bottomrule
\end{tabular}
 }
\label{tab:diff_backbone}
\vspace{-10pt}
\end{table}


\noindent\textbf{Effectiveness of our {\ourMethod} with Different Backbones.} Note that our {\ourMethod} focuses on the design of detection head based on the DETR paradigm. Therefore, to verify the effectiveness of our {\ourMethod}, we decorate our {\ourMethod} detection head with different backbones, including CenterPoint-Pillar~(pillar-based)~\cite{yin2021center}, CenterPoint~(voxel-based)~\cite{yin2021center}, DSVT-Pillar~\cite{wang2023dsvt} and HEDNet~\cite{zhang2024hednet}.
In Table~\ref{tab:diff_backbone}, we present the corresponding detection results on the Waymo validation set~\cite{sun2020scalability} with 20\% training data. We clearly observe that our approach yields consistent performance improvement under different backbones, proving the generality of our SEED detection head.

\noindent\textbf{Effect of $\tau$ for Quality Query Selection.} To explore the effect of the classification score threshold $\tau$ in formula~(4) of the main paper for quality query selection, we set different score thresholds of  $\tau=0.0$, $\tau=0.2$, and $\tau=0.3$, whose results are summarized in Table ~\ref{tab:diff_tau}. When the score threshold is set as 0.0, we find there is a drastic drop in detection performance. Since the predicted object score is close to 0.0, it is more likely to be considered a background object. At this time, the estimated localization scores that are mainly for foreground objects rather than background objects are unreasonable, leading to selecting out poor queries in the stage of quality query selection. 
Therefore, setting a proper score threshold~(\textit{e.g.}, 0.2 or 0.3) to eliminate the negative impact of background objects for quality query selection in DQS is necessary.

\begin{table}[h!]
\vspace{-10pt}
\caption{The effect of different classification score thresholds for quality query selection in dual query selection~(DQS). We use mAP/mAPH~(L2) for evaluating the detection performance.}
\vspace{-10pt}
\small
\centering
\setlength{\tabcolsep}{20pt}
\resizebox{1.0\linewidth}{!}{
\begin{tabular}{l|c|c|c|c}
\toprule
\multicolumn{1}{c|}{\multirow{2}{*}{$\tau$}} & \multicolumn{3}{c|}{3D AP/APH~(L2)} & \multirow{2}{*}{\shortstack{mAP/mAPH \\ (L2)}}\\
& \emph{Vehicle} & \emph{Pedestrian} & \emph{Cyclist}\\
\midrule
0.0    &  66.6/66.2 & 70.1/64.5  & 69.1/67.8  & 68.6/66.2     \\
0.2    &68.5/68.1 & 72.1/66.5 &71.2/70.0  &70.6/68.2    \\
0.3    &  68.7/68.2 & 71.9/66.4  & 71.0/69.8 & 70.5/68.1     \\
\bottomrule
\end{tabular}
 }
\label{tab:diff_tau}
\vspace{-10pt}
\end{table}

\begin{table}[t!]
\vspace{-10pt}
\caption{Effectiveness of our {\ourMethod}. The results are evaluated by the metric of mAP/mAPH~(L1 and L2). We evaluate the latency of our {\ourMethod} for different gird sizes on one NVIDIA GeForce RTX 3090 with a batch size of 1.}
\small
\centering
\setlength{\tabcolsep}{20pt}
\resizebox{1.0\linewidth}{!}{
\begin{tabular}{l|c|c|c}
\toprule
 Grids   &mAP/mAPH~(L1)  &mAP/mAPH~(L2) & Latency~(ms) \\
\midrule
$3\times3$  &  76.7/74.1 & 70.3/67.8  & 73.1    \\
$5\times5$  & 77.0/74.4 & 70.6/68.2 & 74.2    \\
$7\times7$  &  77.1/74.5 & 70.7/68.3  & 77.8 \\
\bottomrule
\end{tabular}
}
\label{tab:ab_girds}
\vspace{-10pt}
\end{table}

\begin{table}[t!]
\caption{The effect of the number of {\ourMethod} decoder layers in transformer decoder for 3D detection performance. We use mAP/mAPH~(L2) to evaluate the detection performance.}
\small
\centering
\setlength{\tabcolsep}{20pt}
\resizebox{1.0\linewidth}{!}{
\begin{tabular}{c|c|c|c|c}
\toprule
\multicolumn{1}{c|}{\multirow{2}{*}{Layers}} & \multicolumn{3}{c|}{3D AP/APH~(L2)} & \multirow{2}{*}{\shortstack{mAP/mAPH \\ (L2)}}\\
& \emph{Vehicle} & \emph{Pedestrian} & \emph{Cyclist}\\
\midrule
1   &  66.8/66.2 & 69.4/62.0  & 69.2/67.8 & 68.5/65.3     \\
3    &  68.4/68.0 & 71.5/65.6 & 70.9/69.7 & 70.3/67.8  \\
6    &  68.5/68.1 & 72.1/66.5 & 71.2/70.0  & 70.6/68.2    \\
\bottomrule
\end{tabular}
 }
\label{tab:diff_layers}
\end{table}

\noindent\textbf{Ablation for Distinct Grids.}  As shown in Table~\ref{tab:ab_girds}, we conduct experiments of varying grid sizes in DGA to investigate their impact on the detection performance and latency. With increasing the grid sizes~($3 \times 3 \rightarrow 5 \times  5$), the detection performance of our {\ourMethod} can be consistently improved in terms of mAP/mAPH~(L2). However, the corresponding computational costs are also increasing due to more sampled features being performed for query interaction, leading to more latency.
Therefore, in our paper, we choose a proper grid size of $5\times5$ as default to trade off the detection performance and latency.

\noindent\textbf{Number of {\ourMethod} Decoder Layers.}  To analyze the effect of different numbers of  {\ourMethod} decoder layers on detection performance, we provide the experimental results in Table~\ref{tab:diff_layers}. When only one {\ourMethod} decoder layer is applied in the transformer decoder to extract contextual features of point clouds, a relatively poor detection performance with 65.3 mAPH/L2 is obtained. In contrast, stacking three {\ourMethod} decoder layers brings an obvious performance gain with 2.5 mAPH/L2 thanks to their more powerful feature extraction capabilities. Intuitively, stacking more {\ourMethod} decoder layers is beneficial. Therefore, in this paper, we adopt the commonly used six {\ourMethod} decoder layers in the transformer decoder, which produces a better result with 68.2 mAPH/L2 than the settings of using fewer decoder layers~(\textit{i.e.}, 65.3 mAPH/L2 for one decoder layer or 67.8 mAPH/L2 for three decoder layers).

\begin{table}[t!]
\small
\caption{The comparison of different query selection in terms of latency.}
\centering
\setlength{\tabcolsep}{20pt}
\resizebox{1.0\linewidth}{!}{
\begin{tabular}{c|c|c}
\toprule
 Query Selection Methods & mAP~/~mAPH~(L2) & Latency~(ms)\\
 \midrule
TransFusion~\cite{bai2022transfusion}~(Heatmap-based) & 67.5~/~65.0 & \textbf{2.5} \\
ConQueR~\cite{zhu2023conquer}~(Top-N) & 69.1~/~66.8 & 7.3  \\
\textbf{{\ourMethod}~(DQS)} & \textbf{70.6~/~68.2} & 10.0 \\
\bottomrule
\end{tabular}
}\label{tab:diff_qs}
\end{table}

\noindent\textbf{Latency of Different Query Selection.}
In Table~\ref{tab:diff_qs}, we provide the latency of different query selection methods. We can observe that our DQS with high performance does not bring significant latency compared with the Top-N method.

\subsection{Different Attention Operations} \label{sec:diff_attn_exp}

\begin{figure}[t!]
\centering
\includegraphics[width=1.0\linewidth]{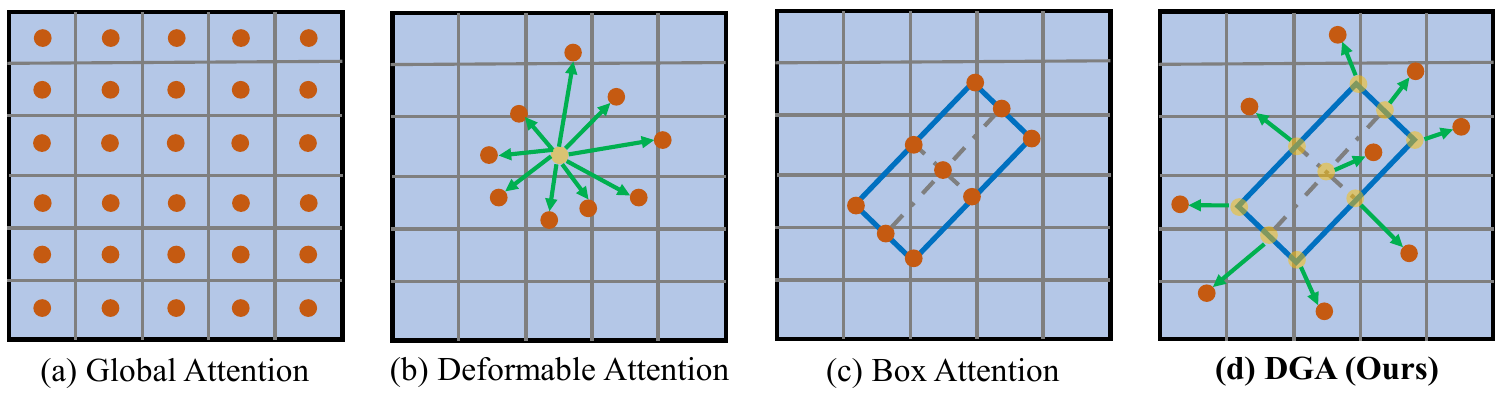}
\caption{Comparison of deformable gird attention~(DGA) with other attention operations. The orange points represent the sampling features, the yellow points represent the reference points, and the green arrows represent the predicted offsets. Note that global attention adopts a global manner for query interaction, that is, treating all features as sampling features.}
\label{fig:supple_dga}
\end{figure}

To clearly illustrate the difference between our proposed deformable grid attention~(DGA) and existing representative attention operations~(\textit{i,e}, global attention~\cite{vaswani2017attention}, deformable attention~\cite{zhu2020deformable} and box attention~\cite{nguyen2022boxer}), we present the simple schematic diagrams of these methods as shown in Figure~\ref{fig:supple_dga}. For the global attention in Figure~\ref{fig:supple_dga}~(a), each query implements feature interaction with all features (as key and value). This operation usually brings unacceptable computational costs, especially for using high-resolution feature maps as keys or values. Therefore, the local attention operations including the deformable attention, the box attention, and our DGA in Figure~\ref{fig:supple_dga}~(b)~(c)~(d)  are more proper to perform query interaction than the global attention in point clouds. Specifically, deformable attention is good at capturing the crucial regions of objects in a flexible receptive filed manner, but the learned offsets without geometric prior information as reference are difficult to predict accurately. The box attention operation can make use of geometric information of some regular objects~(\textit{e.g.}, \emph{Vehicle}), but it requires a precise box regression, and its receptive field is not as flexible as the deformable attention. In contrast, our deformable grid attention has the advantages of both the flexible receptive field of deformable attention and the rich geometric information of the box attention, which can enable the network to focus on relevant regions and capture more informative features even for objects with diverse shapes.

\subsection{Discussion}
\label{sec:discussion}


\begin{table}[h!]
    \centering
    \caption{Comparison of our SEED and FocalFormer3D. $*$ indicates the deformable attention in FocalFormer3D~\cite{chen2023focalformer3d}}
    \vspace{-10pt}
    \subfloat[Comparison for query selection.]{
    \resizebox{0.4\linewidth}{!}{
        \begin{tabular}{c|c|c}
        \toprule
        DQS & multi-stage & mAP/mAPH~(L2) \\
        \midrule
         --& --  & 67.5/65.0 \\
         --& \checkmark  & 68.2/65.5 \\
        \checkmark &-- & 70.6/68.2 \\
        \checkmark & \checkmark & \textbf{70.9/68.3} \\
        \bottomrule
        \end{tabular}
        }
    \label{tab:dqs_ms}
    }
    \subfloat[Comparison for query interaction.]{
    \resizebox{0.5\linewidth}{!}{
        \begin{tabular}{c|c}
        \toprule
        Method  & mAP/mAPH~(L2) \\
        \midrule
        Deformable Attention~\cite{zhu2020deformable}  & 69.9/67.5 \\
        Deformable Attention$*$~\cite{chen2023focalformer3d}  & 70.0/67.6 \\
        DGA~(Ours)  & \textbf{70.6/68.2} \\
        \bottomrule
        \end{tabular}
        }
    \label{tab:dga_da}
    }
\vspace{-25pt}
\end{table}

\noindent \textbf{DQS \textit{vs.} Multi-stages to Select Queries.} Actually, our DQS not only uses a foreground query selection module to select coarse queries with a high recall, but also leverages a quality query selection module to obtain high-quality queries. However, FocalFormer3D~\cite{chen2023focalformer3d} primarily utilizes multi-stage foreground scores to obtain queries with higher recall, but it overlooks the importance of query quality for box localization. Furthermore, we present a comparison between our DQS and the multi-stage approach in Table~\ref{tab:dqs_ms}.
We observe that DQS achieves much better performance~(68.2 \textit{vs.} 65.5), which indicates the importance of selecting high-quality queries. In Table~\ref{tab:dqs_ms}, we also integrate this multi-stage strategy into our DQS, which brings a subtle gain of 0.1 mAPH/L2.

\noindent \textbf{DGA \textit{vs.} Deformable Attention in FocalFormer3D.}
Here, we discuss the difference between our proposed DGA and deformable attention in FocalFormer3D~\cite{chen2023focalformer3d} for query interaction.
In fact, FocalFormer3D adopts the \textbf{same} deformable attention with deformable DETR~\cite{zhu2020deformable}. The only difference with ~\cite{zhu2020deformable} is that FocalFormer3D uses the enhanced queries by combining the RoI features for feature interaction instead of the original queries.  
In contrast, our DGA is a \textbf{new} deformable attention, which uniformly divides each reference box into grids as the reference points and then utilizes the predicted offsets to achieve a flexible receptive field. In Table~\ref{tab:dga_da}, we provide the comparison with FocalFormer3D, whose performance~(67.6 mAPH/L2) is still inferior to our DGA~(68.2 mAPH/L2). Additionally, we provide a clear illustration of the difference between our DGA and deformable attention in Figure~\ref{fig:supple_dga}.

\begin{figure*}[t!]
\centering
\includegraphics[width=0.99\linewidth]{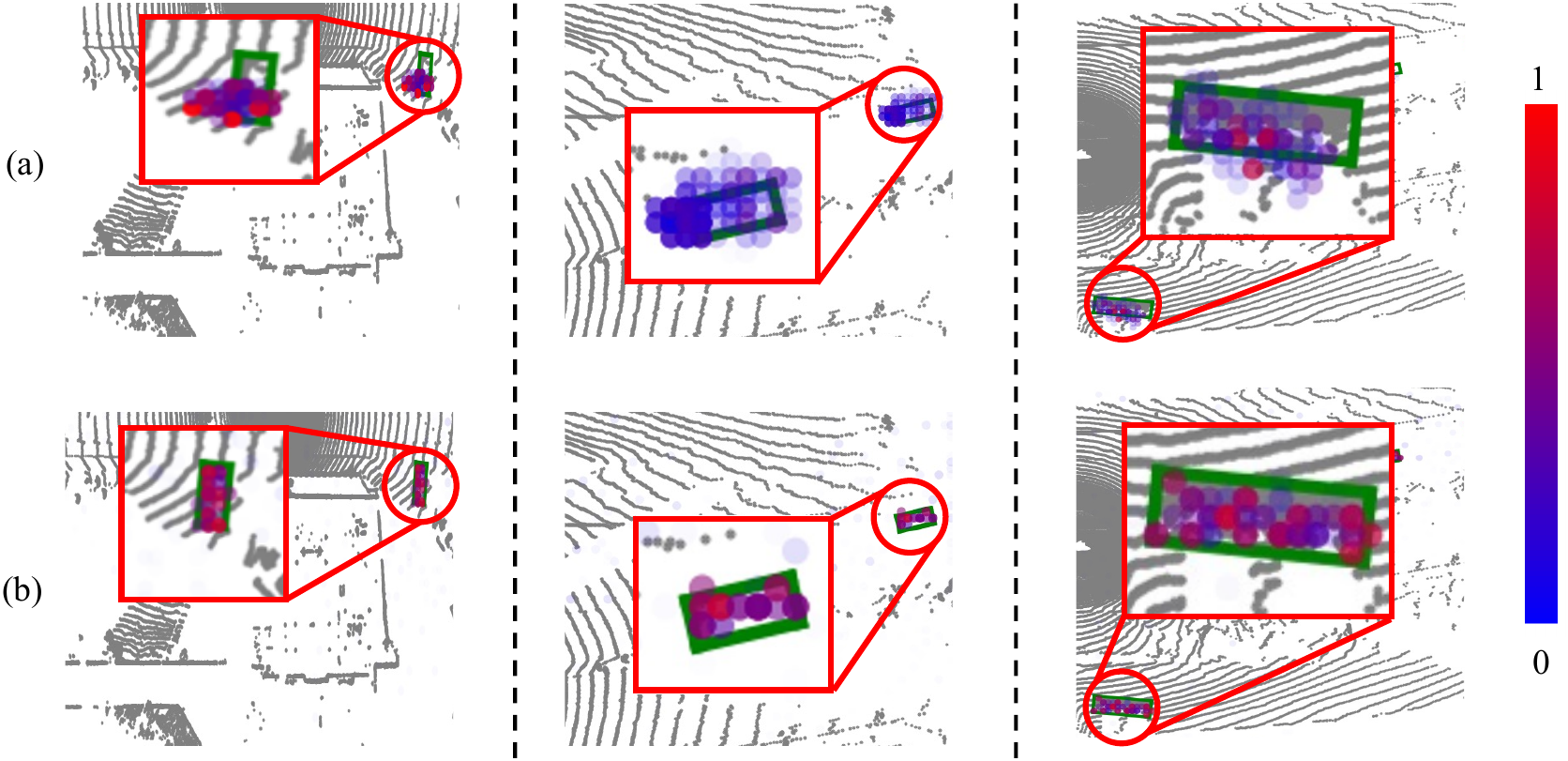}
\caption{Comparison of attention map without DGA (a) and with DGA (b) on the Waymo validation set. Green boxes are the ground truths. The circle represents the position of the attention, and its corresponding color means the weight of the attention.
After utilizing DGA, {\ourMethod} can capture the geometric information of 3D objects in a flexible receptive field and achieve better query interaction.}
\label{vis:dga_att_map}
\vspace{-10pt}
\end{figure*}

\begin{figure*}[t!]
\centering
\includegraphics[width=0.90\linewidth]{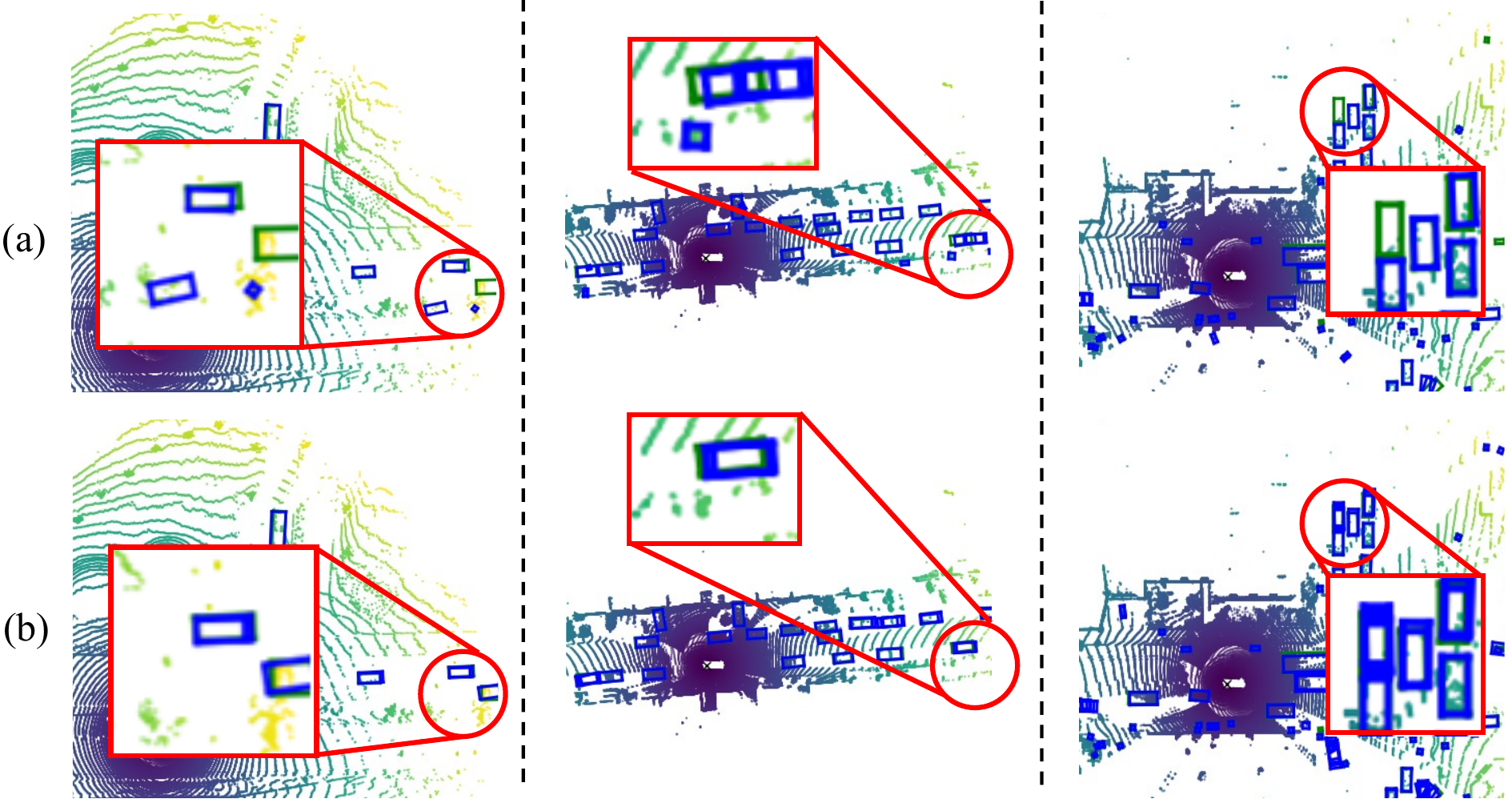}
\caption{Comparison of detection results without DQS (a) and with DQS (b) on the Waymo validation set. Blue and green boxes are the prediction and ground truths, respectively. After utilizing DQS, our {\ourMethod} can successfully detect some hard objects and reduce some false positives, which are highlighted by red circles.}
\label{fig_cmp}
\vspace{-10pt}
\end{figure*}

\begin{figure*}[t!]
\centering
\includegraphics[width=0.9\linewidth]{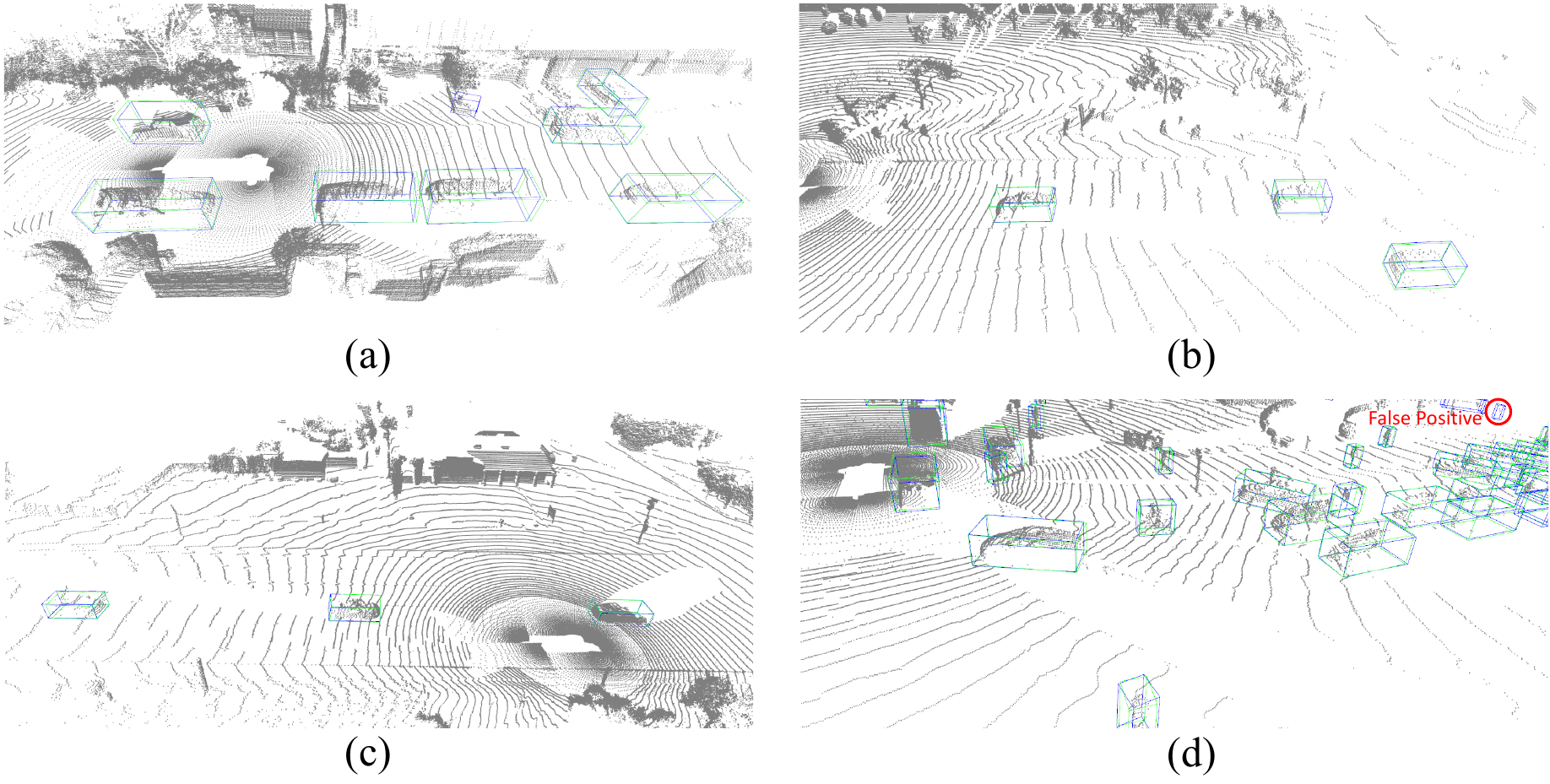}
\caption{Qualitative results of {\ourMethod} on the Waymo validation set. Blue and green boxes are the predictions and ground truths, respectively. Besides, we highlight the false positive with a red circle.}
\label{fig_res}
\vspace{-10pt}
\end{figure*}

\subsection{Visualization} \label{sec:vis}

\noindent\textbf{Visualization of Learned  Attention Map.} As shown in Figure~\ref{vis:dga_att_map}, we present the visualization of learned attention maps under the settings of our {\ourMethod} with DGA~(b) and without DGA~(a)~(\textit{i.e.}, box attention~\cite{nguyen2022boxer}). In the first column, we can observe that DGA captures the key regions even if there is no accurate proposal box as a reference, benefiting from its flexible receptive field. In the second column, we find that DGA produces higher attention weight on objects than the manner without DGA. In the third column, our DGA not only has good robustness in estimating the direction angle but also focuses on key features, such as the boundary and center of the object. The above visualizations effectively demonstrate the superiority of our DGA for query interaction.

\noindent\textbf{Comparisons for w/ and w/o DQS.}
To verify the effectiveness of our DQS, we visualize the detection results of our {\ourMethod} with DQS and without DQS~(\textit{i.e.}, directly select Top $N_f$ queries in one step) on the Waymo validation set, which is depicted in Figure~\ref{fig_cmp}. In the first column, our method can accurately locate all objects and distinguish a False Positive~(FP). Besides, as shown in the second column of Figure~\ref{fig_cmp}, we observe that our {\ourMethod} with DQS can pick out some high-quality queries for accurate localization. Finally, surprisingly, our method has the ability to detect a hard distant object even with some occlusions, as shown in the third column of Figure~\ref{fig_cmp}. These interesting phenomena illustrate the effectiveness of our approach.

\noindent\textbf{Visualization for {\ourMethod}.}
We visualize the qualitative results of {\ourMethod} on the Waymo validation set, which is shown in Figure~\ref{fig_res}. Benefiting from the dual query selection for high-quality query selection and the deformable grid attention for effective query interaction, our {\ourMethod} can detect 3D objects well on large-scale point clouds. Besides, in  Figure~\ref{fig_res}~(d), we carefully find that there are several False Positives~(\textit{e.g.}, \emph{Pedestrian}) in the distant areas. Therefore, we plan to utilize the complementarity of multiple modalities~(\textit{i.e.}, 3D point clouds, and 2D camera images) to distinguish these challenging objects in the future.